%% file: main.tex
\documentclass[letterpaper, 10 pt, conference]{ieeeconf}  

\IEEEoverridecommandlockouts                              

\usepackage{graphicx}
\usepackage{xspace}
\usepackage{tikz}
\usepackage{comment}
\usepackage{color}
\usepackage{lscape}
\usepackage{rotating}
\usepackage[colorlinks]{hyperref}
\usepackage{amsmath,amssymb} 
\usepackage[accsupp]{axessibility}  
\usepackage{arydshln} 
\usepackage{capt-of}

\usepackage{multirow}
\usepackage{booktabs,siunitx,caption} 

\newcommand{\ie}{\textit{i.e.}\xspace}
\newcommand{\eg}{\textit{e.g.}\xspace}
\newcommand{\method}[1]{\textsc{#1}}
\newcommand{\ourmethod}{\method{CAPIT}\xspace}

\newcommand{\cone}[1]{\textcolor{blue}{#1}} 
\newcommand{\ctwo}[1]{\textcolor{purple}{#1}} 
\newcommand{\cthree}[1]{\textcolor{violet}{#1}}
\newcommand{\cfour}[1]{\textcolor{teal}{#1}}

\usepackage{caption}
\usepackage{etoolbox}
\usepackage{adjustbox}
\AtBeginEnvironment{tabular}{\footnotesize}
\title{\LARGE \bf
Image-to-Image Translation for Autonomous Driving from Coarsely-Aligned Image Pairs
}
\author{Youya Xia$^{*\dagger}$, Josephine Monica$^{*\ddag}$, Wei-Lun Chao$^\mathsection$, Bharath Hariharan$^{\dagger}$,\\ Kilian Q Weinberger$^{\dagger}$, Mark Campbell$^{\ddag}$ 
\thanks{$^*$ Equal contributions}
\thanks{$^{\dagger}$ Computer Science Department, Cornell University
        {\tt\small \{yx454, bh497, kqw4\}@cornell.edu}}%
\thanks{$^{\ddag}$ Mechanical and Aerospace Engineering Department, Cornell University
        {\tt\small \{jm2684, mc288\}@cornell.edu}}%
\thanks{$^{\mathsection}$ Department of Computer Science and Engineering, the Ohio State
University 
        {\tt\small  chao.209@osu.edu}}%
}
\begin{document}
\maketitle
\thispagestyle{empty}
\pagestyle{empty}
\input{sec/abstract}
\input{sec/intro}
\input{sec/related}
\input{sec/method}

\input{sec/exp}

\input{sec/conclude}

{\small
\bibliographystyle{unsrt}
\bibliography{ref}
}
\end{document}

%% file: sec/abstract.tex
\begin{abstract}
A self-driving car must be able to reliably handle adverse weather conditions (\eg, snowy) to operate safely.
In this paper, we investigate the idea of turning sensor inputs (\ie, images) captured in an adverse condition into a benign one (\ie, sunny), upon which the downstream tasks (\eg, semantic segmentation) can attain high accuracy. 
Prior work primarily formulates this as an \emph{unpaired} image-to-image translation problem due to the lack of paired images captured under the exact same camera poses and semantic layouts. 
While perfectly-aligned images are not available, one can easily obtain coarsely-paired images. For instance, many people drive the same routes daily in both good and adverse weather; thus, images captured at close-by GPS locations can form a pair.
Though data from repeated traversals are unlikely to capture the same foreground objects, we posit that they provide rich contextual information to supervise the image translation model.
To this end, we propose a novel training objective leveraging \emph{coarsely}-aligned image pairs. 
We show that our coarsely-aligned training scheme leads to a better image translation quality and improved downstream tasks, such as semantic segmentation, monocular depth estimation, and visual localization.
\end{abstract}

%% file: sec/intro.tex
\section{Introduction}
Although the development of autonomous driving perception has rapidly advanced in recent years, work has typically focused on ideal weather conditions. However, to ensure safe operation, these perception systems must also operate robustly under various adverse weather conditions (such as snowy and night)~\cite{pitropov2021canadian}.
One possible solution is to train joint or specialized models to cover all conditions. 
However, this requires having more models or increased model size, as well as costly labeling and retraining for all tasks and conditions. 
Moreover, sensor signals (\eg, images) in adverse weather are likely of poor quality, making some labeling tasks harder. 
In this paper, we investigate an alternative idea of turning images captured in adverse weather into a benign one (\ie, sunny), upon which the downstream tasks are more likely to attain high accuracy, even without additional retraining.

This problem generally can be cast as an image-to-image translation task~\cite{wang2018high}, training neural networks to transfer images from a source domain (\eg, night) to another target domain (\eg, sunny). 
Image-to-image translation training can typically be categorized into paired \cite{isola2017image} or unpaired options \cite{CycleGAN2017,zhao2020aclgan,huang2018munit}. 
Paired translations are trained with perfectly-aligned source and target image pairs, learning to map one to another. While this approach is straightforward, collecting such image pairs is challenging, if not impossible, in many application scenarios. For example, it is infeasible to obtain two driving images with the exact same semantic layouts (\eg, the same set of cars and pedestrians and their placements) under different weather conditions. Thus, for many problem setups (\eg, autonomous driving), existing works usually drop the notion of paired training and utilize the more challenging unpaired training, learning to map between two image distributions, usually accompanied by additional constraints (\eg, cycle consistency~\cite{CycleGAN2017}).

\input{figurestext/introfigure}
In this paper, we propose Coarsely-Aligned Paired Image Translation (\ourmethod), a new fashion for learning the image-to-image translation model by leveraging \emph{coarsely-aligned} image pairs (see \autoref{fig:intro}). 
We argue that \emph{coarsely-aligned} image pairs are much easier to obtain. For instance, many of us drive the same routes daily (\eg, to and from work and school), even under different weather conditions.
We can form coarsely-aligned pairs by picking images that are closest in GPS locations across traversals. While such data may not be perfectly aligned due to the shifts in camera poses or mobile objects at different times, they offer a wealth of training signals.
For example, they reveal consistency and appearance changes of background objects between image domains. Moreover, as we will show, these provide sufficient (weak) supervision to train state-of-the-art image-to-image translation models without perfectly-paired data. 

Our approach addresses inconsistencies across coarsely aligned images in several ways.
First, as two images taken at different times are unlikely to be captured from the exact same camera pose, we propose a misalignment-tolerating L1 loss at the pixel level that first searches for the best local pixel alignment. 
Second, as foreground objects (\eg, vehicles, pedestrians, and cyclists) may vary in terms of their identities and placements, we propose to mask them and apply reconstruction loss only to unmasked regions.
Furthermore, to accommodate stochastic variations (\eg, due to tree swaying or shadow patterns), we propose a novel usage of the noise contrastive estimation (NCE) loss~\cite{park2020contrastive}. 
This type of loss has been used in different contexts (\eg, image retrieval) to overcome significant misalignment effectively.   

We validate \ourmethod on the Ithaca365 dataset~\cite{diaz2022Ithaca365} with driving scenes through the same routes under various weather conditions (\eg, sunny, snowy, and night). 
We show that \ourmethod results in a more faithful image translation model compared to existing paired or unpaired algorithms, not only in the quality of generated images but also in the performance of downstream tasks such as semantic segmentation~\cite{zhao2017pyramid}, depth estimation~\cite{lee2019big}, and visual localization~\cite{naseer2018robust}.

While we focus on autonomous driving applications, our proposed approach is general and applicable to other applications where coarsely-aligned data are easily obtainable. For example, in remote sensing~\cite{campbell2011introduction,dubayah2000lidar}, different sensors (\eg, airborne LiDAR and satellite cameras) capture different channels (\eg, height, infrared, and RGB images) with different characteristics for downstream tasks. While these sensors mainly operate alone, they may have surveyed some common areas at different times. Therefore, we can pair their data by GPS locations to learn to transfer one information channel to the others. 

%% file: figurestext/introfigure.tex
\begin{figure}[t]
    \begin{minipage}{1\linewidth}
        \centering 
        \includegraphics[width=\linewidth]{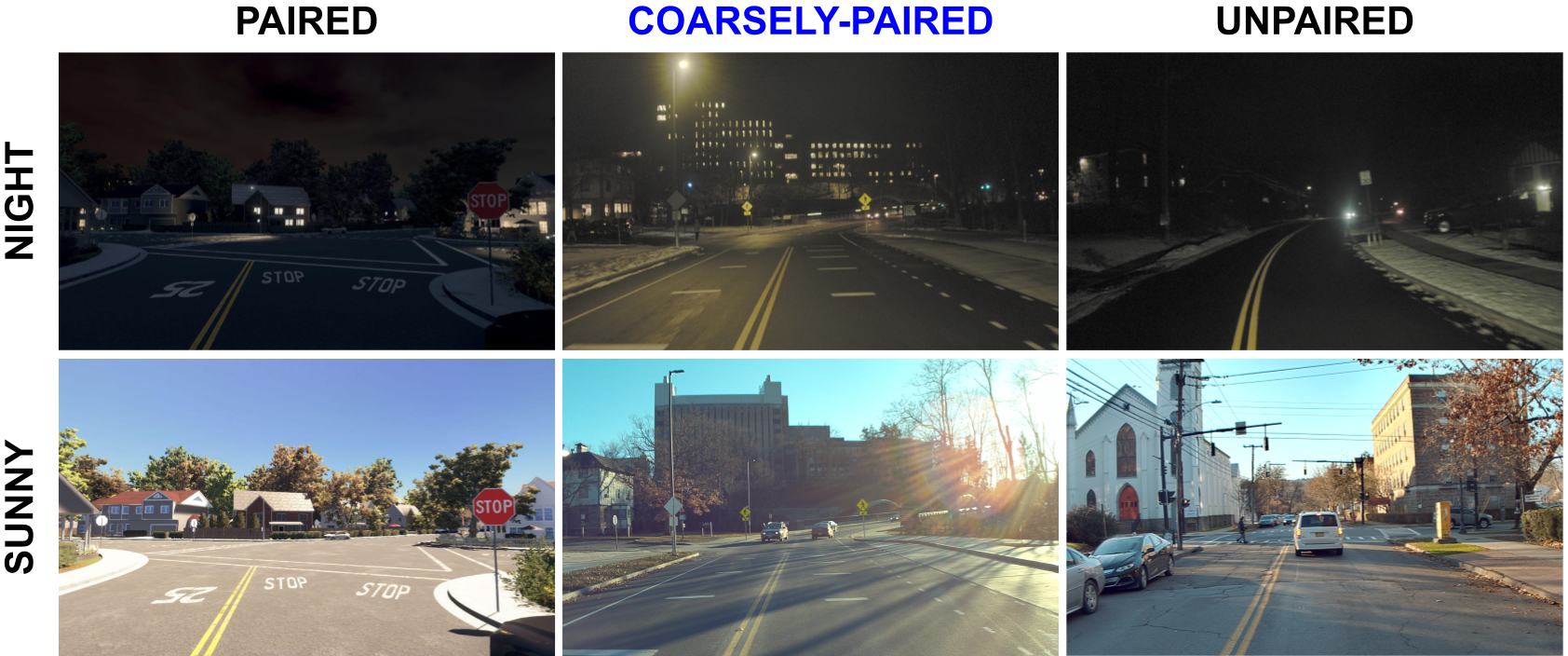}
      \caption{\small Left: perfectly-aligned pair from Apollo \emph{Synthetic} dataset~\cite{huang2018apolloscape}. Mid: coarsely-aligned pair from multiple traversals in Ithaca365 dataset~\cite{diaz2022Ithaca365}. Right: unaligned images~\cite{diaz2022Ithaca365}.\label{fig:intro}}
    \end{minipage}\vskip-15pt
\end{figure}

%% file: sec/related.tex
\section{Related Work}
Early image-to-image translation works focus on paired translation methods, where they apply losses that leverage the strong one-to-one correspondence between input images from the source domain and corresponding images in the target domain to guide the image translation. 
For example, \cite{isola2017image} applies GAN \cite{goodfellow2014generative} in a conditional setting as well as L1 loss between the input and paired target images.
\cite{wang2018high} further adds feature matching and perceptual loss and employs a multi-scale generator and discriminator. 
\cite{Park_2019_CVPR} proposes a spatially-adaptive normalization layer to improve the fidelity and alignment of translated images with the input layout.
\cite{PixelNN} combines GAN with a classic nearest-neighbor approach to generate high resolution image.

However, obtaining perfectly-aligned paired images is often challenging, if not impossible, in many cases (such as autonomous driving). Thus, the loss functions and training framework designed for paired image translation cannot be applied. 
Another line of work focuses on unpaired image translation, where image correspondence is not required. For example, CycleGAN~\cite{CycleGAN2017}, ToDayGAN~\cite{anoosheh2019night}, and DiscoGAN~\cite{kim2017learning} use cycle-consistency loss to enforce consistency between the original image and the reconstructed image after one translation cycle. Council-GAN~\cite{nizan2020council} uses a council loss to achieve similar goal.
While cycle-consistency loss is a widely used constraint, ACL-GAN~\cite{zhao2020aclgan} proposes to use adversarial-consistency loss to encourage the translated image to retain important information from the source image without the need of cycle translation (\ie, translated image translated back to source domain).
CUT~\cite{park2020contrastive} proposes a patchNCE loss that employs a contrastive learning technique~\cite{gutmann2010noise} to maximize mutual information between input and output (translated) images.
Other methods, such as MUNIT~\cite{huang2018munit}, DRIT~\cite{DRIT_plus}, UNIT~\cite{liu2017unsupervised}, and ForkGAN~\cite{zheng2020forkgan} focus on learning domain agnostic features to extract shared features between input and output domains. 

Although unpaired image translation methods allow training without paired images, the unpaired setup leads to an under-constrained problem and inferior results compared to paired methods. 
Our proposed work \ourmethod, Coarsely-Aligned Paired Image Translation, offers an intermediate solution by leveraging \emph{coarsely-aligned} image pairs that are easily obtainable for autonomous driving. To our knowledge, the closest work to our proposed \ourmethod is \cite{porav2018adversarial}.
Our work \ourmethod significantly differs from \cite{porav2018adversarial} in that, we propose a framework dedicated to learning image translation using coarsely-aligned image pairs, while the main training in \cite{porav2018adversarial} comes from the unpaired cycle-consistency training. 
\cite{porav2018adversarial} requires \emph{manually} selecting \emph{well-aligned} images to fine-tune the image translation model after the main unpaired training stage. During fine-tuning, \cite{porav2018adversarial} immediately treats those images with \emph{paired} translation loss and does not address possible misalignment factors, which we show in \autoref{ss_loss} and \autoref{sec:ablation} can adversely degrade the resulting generative model.

While some downstream tasks may require removing only the \emph{scattering medium} (\eg snowfall and raindrop)~\cite{shi2021zeroscatter} from adverse images, adapting whole street scenes (\eg, removing snow from both the street and cars) can generally produce images that are closer to the target domain  and benefit a variety downstream adaptation tasks that may include backgrounds, such as semantic segmentation and depth estimation.
We reiterate that our contribution in leveraging coarsely-aligned image pair translation is \textit{orthogonal} and \textit{complimentary} to existing novel image translation methods and architectures, and can even improve many of them. 
 
Another line of  work is improving downstream tasks (\eg semantic segmentation) in 
adverse weather by directly adapting task networks~\cite{pfeuffer2019robust,naseer2018robust,sharma2020nighttime}; this allows task-specific information to be used to tailor the approach to the task. 
However, this often requires task labels (\eg, semantic mask in the ACDC dataset~\cite{sakaridis2021acdc}), which is often challenging or infeasible to annotate under adverse conditions. Adapting through image translation, on the other hand, does not need additional labels. Moreover, image translation immediately bridges the domain gap for \emph{all} image-based downstream tasks (\eg, semantic segmentation, visual localization). 

%



%% file: sec/method.tex
\section{Proposed Approach: \ourmethod}
\input{figurestext/pipelinefigure}
\subsection{Problem Setup and Baseline Algorithm}
Our goal is to learn an image mapping function $G$ from a source domain $\mathcal{X}$ (\eg, adverse weather) to a target domain $\mathcal{Y}$ (\eg, sunny). 
To begin, we introduce a \emph{paired} translation baseline, inspired by pix2pix~\cite{isola2017image}.
\cite{isola2017image} assumes that the training data are perfectly paired images $\{(x^n,y^n)\}_{n=1}^N$, $x^n,y^n\in \mathbb{R}^{h\times w}$, where $x^n \in \mathcal{X}$ corresponds to $y^n \in \mathcal{Y}$. It introduces an L1 loss comparing translated image $G(x^n)$ to its corresponding target image $y^n$ in the pixel space:
\begin{equation}\label{eq:L1original}
    \mathcal{L}_{L1}(G) = \mathbb{E}_{x^n,y^n}\lVert  G(x^n) - y^n \rVert_1.
\end{equation}
\cite{isola2017image} also employs a conditional GAN loss~\cite{goodfellow2014generative}:
\begin{multline}
    \mathcal{L}_\text{cGAN}(G,D) = \mathbb{E}_{x^n,y^n}\left[\log D(y^n;x^n)\right] +  \\ \mathbb{E}_{x^n}\left[\log \left(1-D(G(x^n);x^n)\right)\right],\label{e_baseGAN}
\end{multline}
where $D$ denotes a discriminator (\ie, binary classifier) that tells a real image $y$ from a fake/generated one $G(x)$; the discriminator $D$ is conditioned on $x$.
Through minimizing both losses with respect to $G$, we encourage the generated images to look real and similar to the target images.

\subsection{Learning with Coarsely-Paired Images}
\label{ss_loss}
Obtaining paired images is often infeasible in many scenarios, including autonomous driving, thus prior works typically drop the notion of paired training and utilize the more challenging unpaired training way. 
Our method leverages \textit{coarsely-aligned} images, which are much easier to obtain. 
Particularly, in autonomous driving, we can obtain \emph{coarse} pairs between $x$ and $y$ if they are collected from the same area, which is common (\eg, driving to work daily).
The Ithaca365 dataset~\cite{diaz2022Ithaca365} collects images and GPS on the same route under different weather conditions (\eg, sunny, night, and snowy). 
We pair images from different traversals  based on the closest GPS pose (see \autoref{fig:intro}-mid for a training data sample). These image pairs are \emph{not perfectly aligned} due to potential GPS differences/errors, changes in semantic layouts at different collection times, and shifts in camera pose. 

Learning with the above two loss functions (\autoref{eq:L1original} and \autoref{e_baseGAN}) does not explicitly account for the misalignment in the coarsely-aligned image pairs and may adversely degrade the resulting generative model. 
Therefore, instead of directly employing the paired translation algorithms~\cite{isola2017image} or entirely ignoring any useful paired information in the training data by applying an unpaired algorithm~\cite{CycleGAN2017,huang2018munit}, we propose \ourmethod, Coarsely-Aligned Paired Image Translation, which is dedicated to learning image translation using \textit{coarsely-aligned} image pairs. Specifically, we propose novel loss functions tailored to coarsely-aligned image pairs to address the aforementioned challenges. \autoref{fig:diagram} shows an overview of our approach.

\subsubsection{Foreground-Masked Loss}
Our first insight about the coarsely-aligned data is that foreground objects (\eg, cars and pedestrians) change in presence and location over time. Thus, two images taken at different timestamps will have different foreground contents and layouts, even if they are taken from the same camera pose. To address this issue, we propose to mask out pixels of the foreground objects in calculating any paired loss. For instance, we calculate the L1 loss only on pixels that belong to the backgrounds in both source and corresponding target images. We denote these background pixel indices as:
\begin{equation}\label{eq:fgmask}
    M(x,y)=\{(i,j) | (c(x_{i,j}) \not\in \text{fg}) \wedge (c(y_{i,j})\not\in \text{fg})\}, 
\end{equation}
where $c(\cdot)$ denotes the class label and fg a set of foreground classes. We can segment these objects using annotated/ground-truth masks or off-the-shelf segmentation models~\cite{he2017mask}.

\subsubsection{Misalignment-Tolerating L1 Loss}
The original L1 loss penalizes differences in pixel values between the generated and target images at each pixel location, assuming that the two images are perfectly aligned. 
However, with coarsely-aligned data, the source and target images are unlikely from the same camera poses. Thus, a conventional L1 loss would exaggerate even with a few pixel shifts between the two images (\eg, along the edges), adversely pushing the generator to fit noises. 
To address this problem, we propose to compare pixel $G(x)_{i,j}$ in the translated image 
to a set of pixels within a small rectangle window around $i,j$. Specifically, we refer to those pixel indices as
\begin{equation}\label{eq:k}
    r(i,j)=\{(a,b):  |a-i|\leq k_h \wedge |b-j|\leq k_w \},
\end{equation}
where $k_h$ and $k_w$ are the size of the rectangular window. Given pixel location $i,j$ in the translated image, we first find (within the search window $r(i,j)$) the \emph{best} pixel location $i',j'$ in the target image to overcome the slight pixel-to-pixel misalignment.
The misalignment-tolerating L1 loss can be formulated as:
\begin{equation}
    \mathcal{L}_{L1}^{\dagger}(G) = 
    \mathbb{E}_{x^n,y^n}\frac{1}{hw}\sum_{i,j}\min_{i',j'\in r(i,j)} | G(x^{n})_{i,j}-y^{n}_{i',j'}|. \label{e_new_L1}
\end{equation}
Combining \autoref{e_new_L1} with the foreground masking defined in \autoref{eq:fgmask}, we obtain the final L1 loss:
\begin{multline}
    \mathcal{L}_{L1}^{\star}(G) = \mathbb{E}_{x^n,y^n} \frac{1}{\text{card}({M(x,y)})} \\
      \sum_{(i,j)\in M(x,y)} \min_{(i',j')\in \{r(i,j)\cap M(x,y)\}} | G(x)_{i,j}-y_{i',j'}|,
\end{multline}
where $\text{card}(\cdot)$ is the cardinality/number of pixels.

\subsubsection{Ranking-based Loss to Overcome Large, Stochastic Variations}
The misalignment-tolerating L1 loss can effectively compensate for small pixel shifts due to slightly different camera poses.
However, two images captured at different times contain inevitable \emph{stochastic} factors that could affect the image contents and cause significant misalignment, even if they are from the same camera pose with masked-out foreground objects.
For example, a tree might be posed differently due to wind, which changes the leaves' position and shadow patterns on sidewalks. 
These stochastic variations are not accounted for by the previous two strategies and could significantly distort the pixel values. 
Thus, we propose a paired loss term that is more lenient to these stochastic variations.
Concretely, we build upon a key insight: although stochastic variations exist, patches of translated and corresponding target images from the same position should still be more similar (\eg, similar structures and contents) than patches from a different position.

We realize this idea by \emph{adapting} the patchNCE loss from \cite{park2020contrastive}. Being an unpaired translation method, 
\cite{park2020contrastive} computes the patchNCE loss between the source and translated images. Unlike \cite{park2020contrastive}, we compute the patchNCE loss between the translated and coarsely-aligned target images, leveraging our coarsely-aligned data.
Let $H$ be a feature extractor that reuses the encoder of the translation model $G$, followed by a two-layer MLP; and $H_s^l$ be the extracted feature at a spatial location $s$ and layer $l$. Our patchNCE loss is:
\begin{multline}
     \mathcal{L}^*_\text{patchNCE}(G,H) = 
     \mathbb{E}_{x^n,y^n} \\ \sum_{l=1}^{L}\sum_s \ell\left(H_s^l(G(x^{n})), H_s^l(y^{n}), \{H_{s'}^l(y^{n}) | s'\neq s\}\right),
\end{multline}
where $\ell$ is a cross-entropy loss to encourage the query $v=H_s^l(G(x^n))$ to match (more similar) with the positive/correct sample $v^\text{+}=H_s^l(y^n)$ over negative samples $\{v^\text{-}\}=\{H_{s'}^l(y^{n}) | s'\neq s\}$:
\begin{equation}
    \ell(v,v^\text{+}, \{v^\text{-}\})=\\
    -\log \frac{\exp(v\cdot v^\text{+}/\tau)}{\exp(v\cdot v^\text{+}/\tau)+\sum_{\{v^\text{-}\}} \exp(v\cdot v^\text{-}/\tau)}.
\end{equation}
Here, $\tau$ is a temperature constant scalar. 
Furthermore, we apply $M(x,y)$ to mask out the foreground objects on the patchNCE computation. 
\subsection{Final \ourmethod Training Objective}
Our final \ourmethod training objective combines the loss terms introduced in \autoref{ss_loss} with an unpaired-GAN (uGAN) loss~\cite{CycleGAN2017}:
\begin{equation}
    \mathcal{L}_\text{CAPIT} = \mathcal{L}_\text{uGAN}(G,D) +  \lambda_1 \mathcal{L}_{L1}^\star(G) + \lambda_2 \mathcal{L}_\text{patchNCE}^*(G,H),
\end{equation}
where $\lambda_1,\lambda_2\in \mathbb{R}^+$ are weighting coefficients. 
While it is natural to follow \cite{isola2017image} to use conditional-GAN (cGAN) loss, we \emph{empirically} find that unpaired-GAN (uGAN) loss \cite{CycleGAN2017} results in better performance (see \autoref{sec:ablation} and \autoref{sec:exp-GANloss}); thus we employ uGAN loss.
We learn $G$ and $H$ to minimize the objective, while $D$ to maximize the objective.
\ourmethod is in theory \emph{agnostic} to different model architectures (\ie, $G$ and $D$). In this paper, we adopt a coarse-to-fine generator and a multi-scale discriminator from \cite{wang2018high}.

%% file: figurestext/pipelinefigure.tex
\begin{figure}[htbp]
        \centering
        \vskip2pt
        \includegraphics[width=1\linewidth]{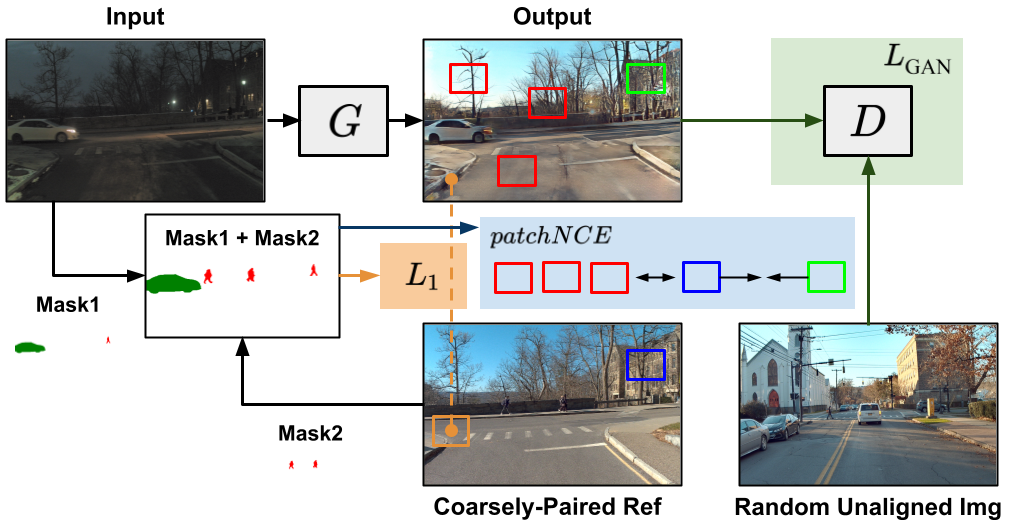}
        \caption{\small \ourmethod framework for training image translation model $G$ using coarsely-aligned image pairs.\label{fig:diagram}}
    \vskip -8pt
\end{figure}

%% file: sec/exp.tex
\section{Experiments}\label{s_exp}
First, we evaluate our \textit{coarsely}-paired image translation approach on the Ithaca365 dataset~\cite{diaz2022Ithaca365} and compare our results to several baseline methods, as shown in \autoref{ss_mainR}. We evaluate the quality of image translation using the FID metric \cite{heusel2017gans} and several downstream tasks. In \autoref{sec:ablation}, we perform an ablation study, demonstrating the key components of our approach in leveraging \emph{coarsely}-paired dataset. In \autoref{sec:exp-foreground}, we provide an analysis on the effect of mask accuracy in foreground-masked loss computation. In \autoref{sec:exp-misalignment}, we provide an analysis on the effect  window size in misalignment-tolerating loss computation. Finally, in \autoref{sec:exp-GANloss} we do an additional experiment on another datasets (Facades \cite{Tylecek13} and map to aerial photos~\cite{isola2017image} datasets) to justify the design choice in our GAN loss.

\subsection{Datasets}
Our main experiment uses the Ithaca365 dataset \cite{diaz2022Ithaca365} which contains diverse scenes (\eg, urban, rural, and campus area) along a repeated 15km route. We use one traversal per weather condition (night and snowy as source domains and sunny as target domain). Each condition is split into $5208/1138$ images for training and testing, where the two splits do not have overlapping location. The images are paired between different conditions based on closest GPS poses. We use pretrained Mask R-CNN~\cite{wu2019detectron2,he2017mask} instance segmentation to obtain the foreground masks for our loss computation in \autoref{eq:fgmask}. 
For additional analysis on GAN loss (\autoref{sec:exp-GANloss}), we also do experiments on additional datasets, Facades \cite{Tylecek13} and map to aerial photos~\cite{isola2017image} datasets.

\subsection{Main Results}\label{ss_mainR}
\begin{figure*}[th]
  \centering 
  \includegraphics[width=1\linewidth]{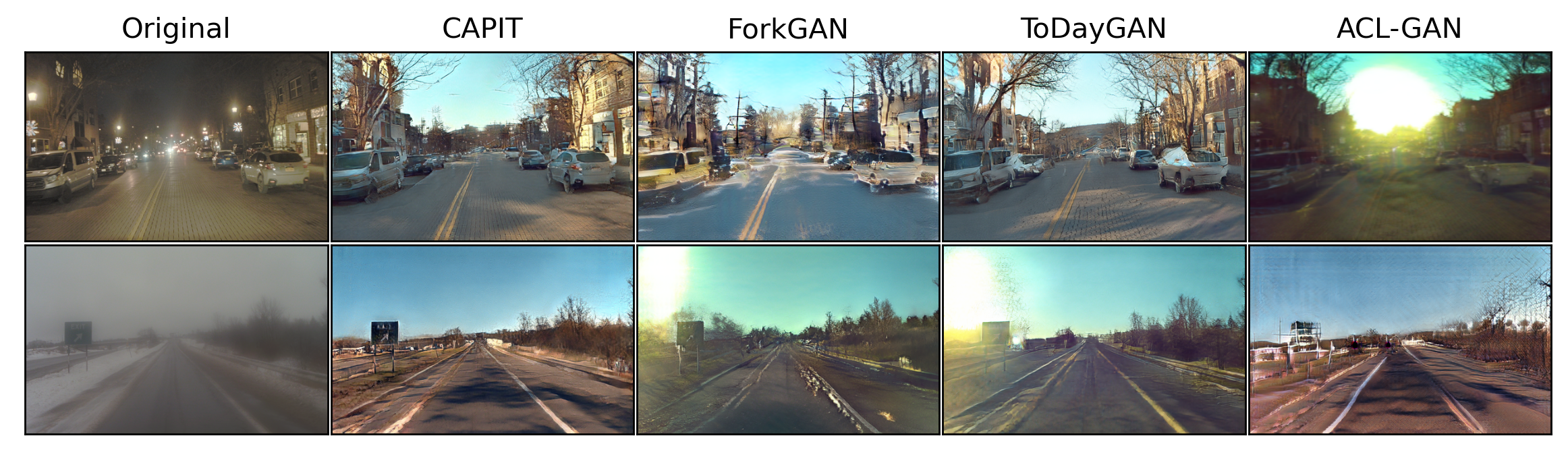}\vskip-5pt
  \caption{\small Visualization results of image translation of our approach compared to several baselines.}
  \label{fig:gen}
  \vskip -15pt
\end{figure*}

We compare our approach to several SOTA unpaired translation models (CUT \cite{park2020contrastive}, ACL-GAN \cite{zhao2020aclgan}, MUNIT \cite{huang2018munit}, CycleGAN \cite{CycleGAN2017}, ForkGAN~\cite{zheng2020forkgan}, and ToDayGAN~\cite{anoosheh2019night}) to show the benefit of leveraging \emph{coarsely}-paired images for supervision. 
\autoref{fig:gen} shows some qualitative results of translated images, which generally demonstrate that our approach can translate an image closer to the reference image.
\autoref{table:comparison-all} shows the performance of our approach compared with other baselines evaluated through different metrics and dowstream tasks (\ie, semantic segmentation, monocular depth estimation, and visual localization).
\subsubsection{FID} We assess the quality of the translated images through Frechet Inception Distance (FID)~\cite{bynagari2019gans} to indicate distance between the distribution of translated images (fake sunny images) and of target images (real sunny images). \autoref{table:comparison-all} shows that our approach has lower FID than the baselines, meaning that our translated images are more similar to the real sunny images.
We also compute the FID between real sunny images and original adverse weather images, \ie, snowy and night, to show the performance gap in adverse conditions without any domain adaptation. Finally, the FID between two different sunny traversals is computed (last row in \autoref{table:comparison-all}) as a lower bound (best case) indication.

\subsubsection{Semantic Segmentation}
We evaluate the performance of image translation model through a downstream task of semantic segmentation using pixel accuracy (PixAcc) as the evaluation metric. 
\emph{All} semantic segmentation experiments are evaluated using the PSPNet~\cite{zhao2017pyramid} pretrained on the MIT ADE20k dataset~\cite{zhou2017scene}. First, we apply the PSPNet model for semantic segmentation directly on night and snowy images. Then, to assess the quality of our image translation model, we apply the same model on the fake sunny images translated from night and snowy using our image translation model, as well as other baseline models. \autoref{table:comparison-all} shows that our image translation method improves the performance of semantic segmentation in adverse conditions more than all baselines.

As the Ithaca365 dataset~\cite{diaz2022Ithaca365} does not provide semantic segmentation labels, we generate \emph{pseudo} ground-truth label by applying the pretrained PSPNet on the sunny \emph{coarse} correspondence images. Additionally, when computing pixel accuracy, if a pixel has a moving foreground label in either the original adverse image or the reference sunny image, it is not included in the accuracy computation. As the pseudo ground-truth reference is approximate, we run semantic segmentation on \emph{another} set of \textit{coarsely-aligned} sunny images, comparing against the pseudo ground-truth to give a sense of achievable upper bound accuracy (last row in \autoref{table:comparison-all}).

\subsubsection{Monocular Depth Estimation}
We also evaluate the performance of image translation through monocular depth estimation downstream task. We use a standard inlier metric $\delta_\text{threshold}$: the percentage of pixels whose inaccuracy level $\max\left(\frac{\hat{d}_{i,j}}{d_i},\frac{d_{i,j}}{\hat{d}_{i,j}}\right)$ are below a threshold (we use 1.25); $d_{i,j}$ and $\hat{d}_{i,j}$ are the ground-truth and predicted depth for pixel $i,j$~\cite{lee2019big}.
We train BTS~\cite{lee2019big} monocular depth estimation model on the sunny data using projected LiDAR as the sparse ground truth. We test the same depth model (trained on sunny condition) on the adverse weather condition images without domain adaptation, as well as on the fake sunny images translated using our approach and several baseline models. Finally, we evaluate the depth model on the in-domain real sunny dataset to give us an upper bound (best case) accuracy (last row in \autoref{table:comparison-all}). We show (\autoref{table:comparison-all}) the benefit of our image translation model compared to the baselines by the domain-induced performance gap that it closes. 
\input{tables/table-comparison}

\subsubsection{Image-based Localization}
\label{ssec:labeling5}
Our final downstream task evaluation is on image-based localization, formulated as an image-retrieval problem. Specifically, given a \textit{query} image, we find the \textit{closest} image using SIFT matching~\cite{lowe1999object} from a set of images in \textit{library}. We evaluate based on localization error (loc-err), \ie Euclidean distance from query image location to its match.
We use images from a sunny traversal as the \textit{library}, while the query images come from the adverse weather images (night and snowy).  First, we directly use the adverse weather image as the \textit{query} without any adaptation; this serves as a bottom baseline. Then, we employ image translation models to first convert the night/ snowy images to fake sunny images for query.
The result in (\autoref{table:comparison-all}) shows that our image translation approach improves the localization error (loc-err) more than the baselines.

\subsection{Ablation Study}\label{sec:ablation}
We perform an ablation study to assess the gains from each of our proposed components and loss terms (shown in \autoref{table:ablation}). 
We begin with a baseline \emph{paired} translation algorithm (Row-1) that employs conditional-GAN (cGAN) and L1 loss~\cite{isola2017image}. 
Comparing Row-1 and Row-2, we see that applying \emph{foreground masking} consistently improves the performance. Comparing Row-2 and Row-3, we see that \emph{unpaired-GAN (uGAN) loss} outperforms the original paired conditional-GAN (cGAN) loss.
Therefore, we employ the uGAN loss instead of cGAN loss. 
Comparing Row-3 and Row-4, we see that using \emph{masked patchNCE loss} between the generated image and the coarsely-aligned target image can largely overcome the stochastic misalignment and improve the performance. Finally, comparing Row-4 and Row-5 shows that accounting for the pixel misalignment in the \emph{misalignment-tolerating L1 loss} computation can further boost the model.
\input{tables/table-ablation}

In short, while we have shown (\autoref{ss_mainR}) the benefit of leveraging the coarsely-aligned data, \emph{the gains are by no means trivial.} Indeed, by comparing Row-1 in \autoref{table:ablation} (baseline paired translation) to existing unpaired algorithms in \autoref{table:comparison-all}, we hardly see the benefit of coarsely-aligned data. However, with our proposed losses, which are dedicated to tackling the misalignment due to foreground objects, camera pose shifts, and stochastic variations, our \ourmethod can leverage the valuable contextual information provided by the coarsely-paired data to excel in image translation.
\subsection{Effect of Mask Accuracy on Foreground-Masked Loss}\label{sec:exp-foreground}
Masks are used to mask out foregrounds (which are unlikely static) in the reconstruction loss computation during \textit{training} only; they can be pre-generated prior to training. As masks are only used for training, \emph{ground-truth masks can be used} if available. If the dataset does not provide such labels (such as in the Ithaca365 dataset), we can can apply pre-trained Mask R-CNN to each \textit{coarsely}-aligned image pair and take the \textit{union} as masks (\autoref{eq:fgmask}). 
\autoref{table:maskthreshold} shows the FID evaluation of image translation different \emph{confidence thresholds} are used for Mask R-CNN mask proposals acceptance. Higher threshold produces lower FP but higher FN. False positive (FP) errors mask out loss computation on unnecessary area, reducing the training signal amount. False negative (FN) errors compute reconstruction loss on foreground areas that may not match in the two images, causing \textit{noise} in the training signal. As shown, Mask R-CNN \emph{default} threshold (0.5) produces the most accurate masks.
\input{tables/maskthreshold}


\subsection{Effect of Window Size on Misalignment-Tolerating Loss}\label{sec:exp-misalignment}
Window size (in \autoref{eq:k}) is studied in \autoref{table:windowsize}.
If images are perfectly aligned, we can directly compare pixels of two images at the same location. However with misalignment, the window acts a search space of correspondence when comparing the reconstruction loss. 
A window too small cannot cover misaligned pixels, while too large over-simplifies the translation problem (\ie, find similar but irrelevant pixels). We use $k_w\!=\!k_h\!=\!k\!=\!3$ in \autoref{eq:k}. 
\input{tables/windowsize}

\subsection{Additional Analysis on GAN Loss}\label{sec:exp-GANloss}
In the previous ablation study, we find that using unpaired-GAN (uGAN) loss results in better performance than conditional-GAN (cGAN) for our \emph{coarsely-paired} training. Inspired by this finding, we do further analysis (\autoref{table:additional}) on different ways of using GAN loss on well-aligned datasets: Facades \cite{Tylecek13} and map-to-aerial photos~\cite{isola2017image}. Specifically, we train the Pix2pix~\cite{isola2017image} paired translation model using three different GAN losses.


Let (realA, realB) be a \textit{paired} batch of images from domain A \& B.
First, in the \emph{conditional-GAN} training,  we use a conditional-discriminator following the original Pix2Pix setup \cite{isola2017image}, where every image fed into the discriminator is conditioned on the input image (realA).
Second, in the \emph{paired-GAN} training, we feed $G$(realA) and realB into the discriminator $D$.
Third, in the \emph{unpaired-GAN} training, we sample \textit{another} batch of images realB' from domain B, and feed $G$(realA) and realB' into $D$.

Interestingly, we find that the unpaired-GAN loss combined with paired L1 results in the best performance for all three experiments. This result not only aligns with our previous finding and supports our GAN loss design, but also suggests that we should reexamine the discriminator design even with perfectly-aligned image pairs.
Note that while the overall training objectives (\ie, expectation over all images) of paired and unpaired GAN are the same, they lead to different training dynamics of $D$ in mini-batch training. We find this important since GAN training iterates between generator and discriminator training and is known to be less stable.
Specifically, we find that \textit{unpaired} training can better capture the overall difference between real and fake images, while \textit{paired} training overly focuses on each paired batch. 
\input{tables/GANloss}

%% file: tables/table-comparison.tex
\setlength{\tabcolsep}{1pt}
\begin{table}[t!]\vskip5pt
   \caption{Evaluation of \ourmethod vs other unpaired translation methods through FID and several downstream tasks. We report the result of \cone{night}/\ctwo{snowy} to sunny translation on the Ithaca365 dataset.}\vspace{2pt} 
    \label{table:comparison-all}
    \begin{tabular}{cccccc}
        \toprule
        No & Method & FID$\downarrow$ & PixAcc (\%)$\uparrow$ & $\delta_{1.25}$ (\%)$\uparrow$ & loc-err(m)$\downarrow$\\ 
        \midrule
        1. & No adaptation &  \cone{162.4} / \ctwo{165.2}  & \cone{50.9} / \ctwo{68.4}    & \cone{70.7}  / \ctwo{74.5} & \cone{22.5} / \ctwo{12.1}\\
        \midrule
        2. & CycleGAN        & \cone{158.3} / \ctwo{136.9} & \cone{52.8} / \ctwo{64.2}    & \cone{78.3} / \ctwo{73.8} & \cone{9.3} / \ctwo{5.7}\\
        3. & MUNIT           & \cone{96.3} / \ctwo{162.4} & \cone{47.4}  / \ctwo{58.8}    & \cone{77.0}  / \ctwo{67.1} & \cone{10.8} / \ctwo{7.9}\\
        4. & CUT             & \cone{94.0} / \ctwo{99.2} & \cone{53.8} / \ctwo{67.8}    & \cone{79.5}   / \ctwo{75.8} & \cone{9.3} / \ctwo{5.0}\\
        5. & ACL-GAN        & \cone{89.4} / \ctwo{101.1} & \cone{48.0}  / \ctwo{68.5}    & \cone{77.0} / \ctwo{73.0} & \cone{11.2} / \ctwo{6.6}\\
        6. & ToDayGAN & \cone{83.2} / \ctwo{87.8} & \cone{57.4} / \ctwo{73.5} & \cone{80.7} / \ctwo{76.4} & \cone{8.9} / \ctwo{3.4}\\
        7. & ForkGAN & \cone{80.9} / \ctwo{101.2} & \cone{54.2} / \ctwo{68.7} & \cone{81.4} / \ctwo{76.1} & \cone{8.8} / \ctwo{5.4}\\ 
        8. & \ourmethod (ours)            & \textbf{\cone{69.4}} / \textbf{\ctwo{76.2}} & \textbf{\cone{61.9}} / \textbf{\ctwo{75.2}}     & \textbf{\cone{83.0}} / \textbf{\ctwo{76.8}} & \textbf{\cone{6.3}} / \textbf{\ctwo{2.9}} \\ 
        \midrule
        9. & Real sunny      & 60.1   & 89.2     & 88.5 & 2.2 \\
        \bottomrule
    \end{tabular}
\vskip -10pt
\end{table}
\setlength{\tabcolsep}{1.4pt}

%% file: tables/table-ablation.tex
\setlength{\tabcolsep}{4pt}
\begin{table}[htbp]
\begin{center}
\caption{Ablation study of loss terms for \cone{night}/\ctwo{snowy} to sunny translation on the Ithaca365 dataset.\label{table:ablation}}
\begin{tabular}{lc}
\toprule
Loss & FID $\downarrow$ \\
\midrule
cGAN + L1        & \cone{114.0} / \ctwo{122.1} \\ 
cGAN + L1 (+mask)         & \cone{106.9} / \ctwo{119.8}  \\
uGAN + L1 (+mask)    & \cone{94.8} / \ctwo{114.8}\\
uGAN + L1 (+mask) + NCE (+mask)   & \cone{68.5} / \ctwo{86.6}\\
uGAN + L1 (+mask + misalignment) + NCE (+mask)        & \cone{69.4} / \ctwo{76.2} \\
\bottomrule
\end{tabular}
\end{center}
\vskip -5pt
\end{table}
\setlength{\tabcolsep}{1.4pt}

%% file: tables/maskthreshold.tex
\begin{table}[htbp]
\begin{center}\caption{Mask confidence threshold study for \cone{night}/\ctwo{snowy} to sunny translation on the Ithaca365 dataset.}\vspace{4pt}
\label{table:maskthreshold}
\begin{tabular}{cccccc}
\toprule
Threshold & 0.9 & 0.7 & 0.5 & 0.3 & 0.1 \\
\midrule
FID $\downarrow$       & \cone{85.6} / \ctwo{92.1} & \cone{79.4} / \ctwo{85.2} & \cone{69.4} / \ctwo{76.2} & \cone{73.2} / \ctwo{80.5} & \cone{81.8} / \ctwo{90.1} \\
\bottomrule
\end{tabular}
\end{center}
\vskip -10pt
\end{table}

%% file: tables/windowsize.tex
\begin{table}[htbp]
\begin{center}\caption{Window size study for \cone{night}/\ctwo{snowy} to sunny translation on the Ithaca365 dataset.}\vspace{4pt}
\label{table:windowsize}
\begin{tabular}{cccc}
\toprule
$k$ & 1 & 3 & 5 \\
\midrule
FID $\downarrow$      & \cone{75.9} / \ctwo{89.7} & \cone{69.4} / \ctwo{76.2} & \cone{73.2} / \ctwo{84.3}\\
\bottomrule
\end{tabular}
\end{center}
\vskip -10pt
\end{table}

%% file: tables/GANloss.tex
\begin{table}[htbp]
\begin{center}\caption{Analysis on three different ways of feeding GAN discriminator loss. \cthree{FID-1} is the FID for \cthree{Label$\rightarrow$Facades}, \cthree{PixAcc-1} is the per-pixel label accuracy for \cthree{Facades$\rightarrow$Label}, and \cfour{FID-2} is the FID for \cfour{Map$\rightarrow$Aerial}.} \vspace{4pt}
\label{table:additional}
\begin{tabular}{lccc}
\toprule
Loss  &  \cthree{PixAcc-1 (\%) $\uparrow$} &  \cthree{FID-1 $\downarrow$} &  \cfour{FID-2 $\downarrow$}\\
\midrule
conditional-GAN + L1    & \cthree{49.2} & \cthree{165.4} & \cfour{111.1} \\
paired-GAN + L1         & \cthree{51.1} & \cthree{130.5} & \cfour{101.2}\\
unpaired-GAN + L1     & \cthree{\textbf{74.6}} & \cthree{\textbf{124.7}} & \cfour{\textbf{91.4}}\\
\bottomrule
\end{tabular}
\end{center}
\vskip-15pt
\end{table}

%% file: sec/conclude.tex
\section{Conclusion}
We propose an approach to leverage the \textit{coarsely-aligned} datasets for image-to-image translation. Our proposed approach, Coarsely-Aligned Paired Image Translation (\ourmethod), is built upon several key insights of the coarsely-aligned data. 
These include foreground masking, local re-alignment to overcome pixel shifts, and a ranking-based loss to address stochastic variation. 
\ourmethod improves both the generated image quality and the performance of downstream tasks in autonomous driving.
Furthermore, our approach is general and can be extended to other applications where coarsely-aligned data is easily available.